\documentclass[a4paper]{article}

\pdfoutput=1

\usepackage{times}

\usepackage[utf8]{inputenc}

\usepackage[cmex10]{amsmath}
\usepackage{booktabs}
\usepackage{amssymb}
\usepackage{enumerate}
\usepackage{multirow}
\usepackage{combinedgraphics}
\usepackage{subfig}

\usepackage[detect-all]{siunitx}

\usepackage{natbib}

\usepackage[commentsnumbered,linesnumbered,ruled]{algorithm2e}

\usepackage{array}
\newcolumntype{L}[1]{>{\raggedright\let\newline\\\arraybackslash\hspace{0pt}}p{#1}}
\newcolumntype{C}[1]{>{\centering\let\newline\\\arraybackslash\hspace{0pt}}p{#1}}
\newcolumntype{R}[1]{>{\raggedleft\let\newline\\\arraybackslash\hspace{0pt}}p{#1}}

\newtheorem{Defn}{Definition}

\newcommand\blfootnote[1]{%
  \begingroup
  \renewcommand\thefootnote{}\footnote{#1}%
  \addtocounter{footnote}{-1}%
  \endgroup
}

\title{A simple efficient density estimator that enables fast systematic search}

\author{Jonathan R. Wells and Kai Ming Ting \\	School of Engineering and Information Technology\\ Federation University Australia.\\		Email: \{j.wells, kaiming.ting\}@federation.edu.au }

\begin{document}

\maketitle

\begin{abstract}
This paper introduces a simple and efficient density estimator that enables fast systematic search. To show its advantage over commonly used kernel density estimator, we apply it to outlying aspects mining.
Outlying aspects mining discovers feature subsets (or subspaces) that describe how a query stand out from a given dataset. The task demands a systematic search of subspaces. We identify that existing outlying aspects miners are restricted to datasets with small data size and dimensions because they employ kernel density estimator, which is computationally expensive, for subspace assessments.
 We show that a recent outlying aspects miner  can run orders of magnitude faster by simply replacing its density estimator with the proposed density estimator, enabling it to deal with large datasets with thousands of dimensions that would otherwise be impossible.
\end{abstract}

\section{Introduction}

%
%

%
%
%
%

%

%

%


%

%

%

%
%
%
%
%
%

Kernel density estimator (KDE)
\citep{Parzen1962,Silverman} has been the preferred choice for many applications \blfootnote{This paper is under consideration at Pattern Recognition Letters.}.  
For example, existing outlying aspects miners such as OAMiner \citep{Duan2015} and Beam \citep{Vinh2016} have employed KDE to estimate density (as the basis) to assess subspaces.

However, the use of KDE has made both OAMiner and Beam computationally expensive. Thus, they cannot be applied to large datasets with thousands of dimensions. 

Grid-based density estimator \citep{Silverman} is often overlooked because it is considered to be an inferior estimator compared to KDE, though it is more efficient.

Here we show that there is a smoothed version of grid-based density estimator which can produce accurate estimation for outlying aspects mining, and at the same time maintaining the efficiency advantage.

We contribute to
\begin{enumerate}
\item An efficient density estimator which runs orders of magnitude faster than kernel density estimator; yet both estimators produce estimations of similar accuracy for the purpose of outlying aspects mining.
\item Enabling an existing systematic search method for outlying aspects mining to  handle large datasets with thousands of dimensions that would otherwise be infeasible. 
\item Verification of the efficiency and effectiveness of the proposed density estimator using real-world and synthetic datasets. We show that it enables an existing outlying miner to run at least two orders of magnitude faster.
\end{enumerate}

The rest of the paper is organised as follows. Related works are provided in Section \ref{sec_related_work}.
 The new simple density estimator, which enables existing outlying aspects miners to run orders of magnitude faster, is described in Section \ref{sec_NeighbourFunction}. The details of employing the density estimator in a recent outlying aspects miner is provided in Section \ref{sec_definitions}.
Section \ref{sec:complexities} explains the time and space complexities. 
Section \ref{sec:Evaluation} reports the empirical evaluation results. 
Conclusions are provided in the last section.

\section{Related works}
\label{sec_related_work}

Kernel density estimator \citep{Silverman,ScottBook} is defined as follows:

\begin{equation}
    f_h(q) = \frac{1}{nh}\sum\limits_{i=1}^nK \left( \frac{q - x^i}{h} \right)
\end{equation}
\noindent
where $K(\cdot)$ is a kernel, $h$ is the bandwidth, $q$ is the query point, and  $D=\{x^i\ |\ i=1,\dots,n\}$ is the given data set. 

Both recent outlying aspects miners Beam \citep{Vinh2016} and OAMiner \citep{Duan2015} used Härdle's rule of thumb \citep{hardle1991} (which is based on Silverman's rule of thumb \citep{Silverman}) to define the bandwidth of KDE. It is given as follows:

\begin{equation}
    h = 1.06\ \min\left\{ \sigma, \frac{IQR}{1.34}\right\} n^{-\frac{1}{5}}
\end{equation}
\noindent
where $IQR=X_{[0.75n]} - X_{[0.25]}$, the interquartile range of the first and third quartiles. 

To compute the density of a query $q$ in a multi-dimensional subspace $s$, a product kernel \citep{Hardle2004} is used:

\begin{equation}
    f_h(q) = {\frac{1}{n\prod\limits_{a \in s} {h_{a}}}} \sum\limits_{i=1}^n \left\{ \prod\limits_{a \in s} K \left( \frac{{q_a} - x^i_a}{{h_{a}}} \right) \right\}
\end{equation}
\noindent
where $h_a$ is the bandwidth in dimension $a$.

Histogram \citep{Silverman,ScottBook} is also well-studied. One robust way to set the bin width is the 
Freedman–Diaconis rule \citep{Freedman1981}: 

\begin{equation}
\mbox{Bin width} = 2 \frac{IQR}{n^{1/3}} 
\end{equation}
\noindent
where $IQR$ is the interquartile range as defined above.

One common way to get smoothed estimations from histogram is average shifted histogram \citep{ASH} or its randomised version \citep{DEMass,RASH,RS-Forest}, where $m$ histograms  having $m$ different shifted origins are built; and each estimation is an average over the $m$ histograms. 
However, $m$ needs to be set to a high number for these methods to work, typically 200 \citep{RASH} or 1000 \citep{DEMass}. This kind of estimators will cost $m$ times that of a single histogram. Thus, it is not suitable for subspace search applications.

A multi-dimensional histogram is constructed using a multi-dimensional grid, where the bin width in each dimension is set based on one-dimensional histogram.

While histogram is more efficient than KDE, it has larger estimation errors than KDE. Though existing smoothed histograms are competitive, they are not significantly faster than KDE when applied to systematic search. This is the reason why many applications have used KDE. Recent works on outlying aspects mining described below are an example.

OAMiner \citep{Duan2015} employs a depth-first search that utilises some anti-monotonicity criteria to prune its search space. Given a query and a dataset, it uses a rank score which ranks instances within each subspace based on density of the instances in the subspace. After exploring all the subspaces (bar pruned subspaces), it selects the (minimal) subspaces in which the query is top-ranked, as the outlying aspects of the query with respect to the dataset.

Vinh et al \citep{Vinh2016} investigate a suite of score functions, and advocate functions which are dimensionality unbiased, i.e., `the average value for any data sample drawn from a uniform distribution is a quantity independent of the dimension of the subspace'. Density is a dimensionality biased score because the density reduced substantially as the number of dimensions increases. They propose to use Z-score as a generic means to convert any dimensionality biased scores to ones which are unbiased. Their evaluation on several scores on some datasets shows that density Z-score is the best. This evaluation is based on a beam search method.

Other existing systematic search methods, such as HiCS \citep{HiCS}, find a list of high contrast subspaces in a preprocessing step, and then apply an anomaly detector to find anomalies for each high contrast subspace. The high contrast score requires a series of statistical tests and a Monte Carlo estimation. It is thus a computationally expensive process, and it needs to be repeated for every candidate subspace, generated from an Apriori-like mechanism as in OAMiner. To improve its runtime, a heuristic pruning rule and an alternative calculation using cumulative entropy must be used \citep{CMI}. Even with this improvement, it is still computationally more expensive than OAMiner \citep{Duan2015} and Beam \citep{Vinh2016}.

The above studies in outlying aspect mining have used computationally expensive estimators such as KDE. This constrains the algorithms' applicability to datasets with low dimensions and small data sizes. We investigate an accurate density estimator based on grid which can expand the algorithms' applicability to large datasets with  high dimensions. %
 
\section{{\bf sGrid}: A Simple Neighbourhood Function for fast density estimation}
\label{sec_NeighbourFunction}

Let $R \in \mathbb{R}^d$ be a $d$-dimensional space of real domain; and $s \subset R$ be a subspace of $k=|s|$ dimensions, where $k \le d$. 

Given a dataset $D$ in $R$, let $x.s$ be a projection in $s$ of an instance $x \in D$; and 
let $N_s(x) = \{ y \in D\; |\; y.s = x.s \}$ be the $s$-subspace neighbourhood of $x$.

To facilitate the estimation of $|N_s(x)|$, we employ $b_z$ bins of equal-width\footnote{The Freedman–Diaconis rule \citep{Freedman1981} mentioned in Section \ref{sec_related_work} can be used to set the bin width and the bin number for each dimension automatically for a given dataset.} for dimension $z$,
and the $s$-subspace neighbourhood consists of the bin covering $x$ and its neighbouring bins in all directions in the subspace. 

Let $\mathfrak{b}_{(i,j)}$ be the bin which covers $x$ and has indices $(i,j) \in \{(1,1),(1,2),\dots, (b_1,b_2-1),(b_1,b_2)\}$ in their respective dimensions in a 2-dimensional space.
The neighbouring bins of $\mathfrak{b}_{(i,j)}$ are $\mathfrak{b}_{(i\pm1,j\pm1)}$.  Neighbouring bins in a subspace having $k$ dimensions are defined similarly with $k$ indices.

\begin{figure}
    \centering
    \includegraphics[scale=0.5]{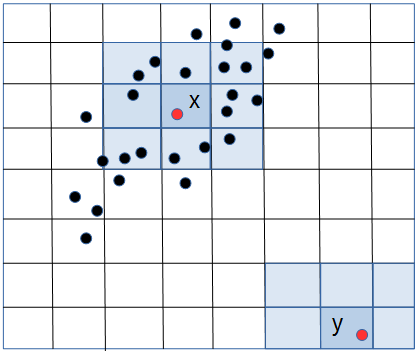}
    \caption{An illustrative example of the new neighbourhood function. The two shaded regions are the neighbouring bins used to estimate density for two instances $x$ and $y$.}
    \label{fig:Example}
\end{figure}

Let $B(N_s(\cdot))$ be the number of bins in $N_s(\cdot)$. The number of bins used to estimate $|N_s(x)|$ is  $B(N_s(x)) = 3^k$ if $x$ falls in a bin which is surrounded by neighbouring bins in all directions; and $2^k \le B(N_s(x)) < 3^k$ when $x$ falls in a bin at the edge of any one dimension; and $B(N_s(x)) = 2^k$ if $x$ falls in a bin at the edges of all dimensions (i.e., at one corner of the subspace). Examples of the estimations using neighbouring bins are shown in Figure \ref{fig:Example}, where $x$ is estimated using 9 bins and $y$ is estimated using 6 bins because the latter is at the edge on one dimension. 

For all $x \ne y$ which are close to each other such that $N_s(x) \cap N_s(y) \ne \emptyset$, the maximum number of bins that overlap between $N_s(x)$ and $N_s(y)$ is $2 \times 3^{k-1}$. 

In a nutshell, the estimation of $|N_s(\cdot)|$ is a smoothed version of that provided by an ordinary grid-based estimation method  which does not employ neighbouring bins (where $B(N_s(\cdot))=1$). We name it: {\bf sGrid} density estimator.

Why does the neighbourhood need to be defined using $B(N_s(\cdot))>1$? In addition to the smoothing effect,
 the ordinary grid-based estimation method  often creates a situation where $N_s(\cdot)$ contains the same small number of instances in both dense and sparse regions, hampering the capability to find subspaces for an anomaly. Inclusion of neighbouring bins in the estimation enables a better distinction between dense and sparse regions. This is depicted in Figure \ref{fig:Example}, where either $x$ or $y$ would have been estimated using one bin, each having 1 instance only, if the ordinary grid-based estimation method is employed. Using the proposed neighbourhood estimation, $x$ and $y$ are clearly differentiated to be in the dense and sparse regions, respectively, because the former is estimated to have 17 instances and the latter has 1 instance.

\noindent
{\bf Cost of computing  $|N_s(x)|$}. The computation can be done through a bitset intersection operation, i.e., $\mathfrak{b}_{(i,j)} = \mathfrak{b}_{(i,.)} \cap \mathfrak{b}_{(.,j)}$. In other words, it takes $k$ intersection operations to get all the instances in the center bin of $N_s(x)$ which has $k=|s|$ dimensions. As there are $3^k$ bins in $N_s(x)$, the total time cost is $O(nk3^k)$. Using a block size $w$ in the bitset operation, the time cost can be reduced to $O(nk3^k/w)$.

In addition, the cost can be significantly reduced by aggregating the three neighbouring  bins, i.e., $\mathfrak{b}_{(i,.)}$ and $\mathfrak{b}_{(i\pm1,.)}$, in each dimension into one pseudo-bin $\mathsf{b}_{(i,.)}$ in preprocessing. A dimension having $b$ bins gives rise to $b$ pseudo-bins. Instead of spending $O(nk3^k/w)$ on $3^k$ bins, $|N_s(x)|$ can now be computed on $k$ pseudo-bins which take $O(nk/w)$ only. This is the same time cost for the ordinary grid-based method (or histogram). 
$w=64$ is used in the experiments reported in Section \ref{sec:Evaluation}.

\section{Using {\bf sGrid} in outlying aspects miner}
\label{sec_definitions}

To examine the capability of the proposed density estimator, we use a recent outlying aspects miner \citep{Vinh2016}. 

We describe the subspace score and the systematic search method used in the following paragraphs.

Density Z-score is identified to be the best score for outlying aspects mining \citep{Vinh2016}. 
The following definition is based on the density estimator stated in the last section.

\vspace{6pt}
\noindent
{\bf Density Z-score}. The definition based on density Z-score is given as follows, when density is estimated using $|N_s(q)|$:

\vspace{2mm}
\begin{Defn}
\bf{A query $q$, which stands out from $D$, has its uniqueness described by any one of the subspaces $s \in S$ in which $q$ is located in a scarcely populated neighbourhood if every subspace $s$ in $S$ has the Z-score normalised number of instances in the $s$-subspace neighbourhood of $q$ less than $\tau$, i.e., $\forall  s \in S, \; Z_{s}(q) < \tau$,
where $Z_s(q)= \frac{|N_s(q)| - \mu}{\sigma}$ is the Z-score\footnote{Z-score is used instead of $|N_s(q)|$ because the latter is a dimensionality biased score, and the former is a dimensionality unbiased score \citep{Vinh2016}. A comparison between subspaces using $|N_s(\cdot)|$ (or density) is biased towards lower subspaces because as the number of attributes increases, the total space/volume increases exponentially wrt the number of attributes.} of $|N_s(q)|$; $\mu$ and $\sigma$ are mean and standard deviation, respectively, of $|N_s(\cdot)|$ in the given dataset.
}
\label{Def_outlying}
\end{Defn}

\vspace{6pt}
Z-Score standardises any measure. For example, a Z-score value of -2 for an instance $q$ means that $q$ is two standard deviations smaller than the mean. For anomalies, we are only interested in bins having small number of instances, i.e., having negative Z-score values.

Vinh et al \citep{Vinh2016} suggest the use of beam search 
to mine outlying aspects. Beam \citep{Vinh2016} employs the standard breadth-first search procedure \citep{RussellAIBook} that explores a fixed number of subspaces (called beam width) at each level in the search process. A minor adjustment is made to explore all subspaces with two attributes \citep{Vinh2016}. The procedure is reproduced in Algorithm \ref{alg_Beam} for ease of reference. 

Using synthetic datasets, Vinh et al \citep{Vinh2016} have shown that Beam performed the best using density Z-score. We use this as the basis for comparison. 

We produce two versions of Beam, denoted as sGBeam and GBeam. sGBeam employs the proposed {\bf sGrid} density estimator  (where $B(N_s(\cdot))>1$) to replace KDE to compute the density Z-score in line 10 of Algorithm \ref{alg_Beam}. GBeam uses the ordinary grid-based density estimator (where $B(N_s(\cdot))=1$). The rest of the procedure is identical.

\begin{algorithm}
 \SetKwInOut{Input}{input}
 \SetKwInOut{Output}{output}

 \Input{$D$ - given dataset, $q$ - query, $\ell$ - maximum search depth, $W$ - beam width, $k$ - top $k$ subspaces as the final output, $A$ - set of attributes defining $D$}
 \Output{$S$ - Set of subspaces for $q$}
 \BlankLine

Initialise $S = \emptyset$\;
Generate all subspaces of one \& two attributes\;
Add the top $k$ subspace to $S$\;
Initialise $L_{(2)}$ with top $W$ subspaces\;

 \For{l=3 to $\ell$}{
  Initialise $L_{(l)} = \emptyset$\;
  \For{each subspace $s \in L_{(l - 1)}$}{
   \For{each attribute $A_i$}{
    \If{\{$s \cup A_i$\} has not been considered yet}{
     Compute density Z-score for \{$s \cup A_i$\}\;
     If the worst scored subspace in $S$ is worse than \{$s \cup A_i$\} then replace\;
     If $|L_{(l)}| < W$ then append \{$s \cup A_i$\}; otherwise, if the worst scored subspace in $L_{(l)}$ is worst than \{$s \cup A_i$\} then replace\;
    }
   }
  }
 }
 \Return $S$\;
 \caption{Beam($q,\ell,D,W,k,A$)}
 \label{alg_Beam}
 \hspace{-10mm}
\end{algorithm}

\begin{table}[h]
 \centering
  \caption{Time cost. %
  $k$ is the attribute size in a subspace. $w$ is the block size in the bitset operation described in Section \ref{sec_NeighbourFunction}. }
  \label{tab:complexity}
 \begin{tabular}{l@{\ \  }l@{\ \ }l}
  \toprule
Method & One estimation & $n$ estimations  \\
  \midrule
  $|N_s(\cdot)|$ & $O(nk/w)$ & --- \\
    $|N_s(\cdot)|$ Z-score  & $O(nk/w)$ & $O(n^2k/w)$ \\
    Density & $O(nk)$ & --- \\
  Density Z-score   & $O(nk)$ & $O(n^2k)$ \\
  \bottomrule
 \end{tabular}
\end{table}

\section{Time and space complexities}
\label{sec:complexities}

The most expensive part in an outlying aspects miner is the score calculation because it needs to be repeated for every instance in each subspace.
The time cost for four different scores are summarised in Table \ref{tab:complexity}. 

Note that Z-score is a secondary score derived from a base score such as density or $|N_s(\cdot)|$. 

Using an efficient method to compute standard deviation (in order to compute density Z-score) takes $O(nk)$ only;
and it only needs to be computed once in each subspace. Therefore, the total time cost of computing density Z-score is $O(n^2k)$ for $n$ instances, which is repeated for each subspace.

In summary, 
although density or $|N_s(\cdot)|$ has the same time complexity, the large constant associated with a density estimator such as KDE makes a huge difference in terms of real compute time. This will be shown in Section \ref{sec:runtime-comparison}.

The algorithm that uses $|N_s(\cdot)|$ requires to store $n$ instances of $d$ attributes and an average of $b$ bins per attribute. Thus, it takes $O((n+b)d)$ space complexity. 
The algorithm that uses KDE also has space complexity linear to $nd$.

\section{Empirical Evaluation}
\label{sec:Evaluation}

The outlying aspects miners used in the comparison are OAMiner \citep{Duan2015}, Beam \citep{Vinh2016}, sGBeam and GBeam. OAMiner and Beam are the latest state-of-the-art outlying aspects miners that use KDE. We use the OAMiner code made available by the authors \citep{Duan2015}.   
All algorithms  are implemented in JAVA\footnote{Note that we use the JAVA implementation of Beam that employs KDE in the experiments 
because the MATLAB implementation provided by the authors \citep{Vinh2016} produces worse results (see details in Section \ref{sec:OAMinerDatasets}).}. The only difference between Beam, sGBeam and GBeam, which directly influences their runtimes, is the density estimator used.

The default settings used for these algorithms are: OAMiner ($\alpha=1$), all beam search methods  use beam width = 100 (as used in \citep{Vinh2016}). The $N_s(\cdot)$ density estimator employed in sGBeam and GBeam uses %
equal-width bins per dimension; and the  Freedman–Diaconis rule \citep{Freedman1981} is used to set the bin width \footnote{We have attempted other rules, such as Sturges rule \citep{Sturges} and Scott's rule \citep{Scott1979}, to set the bin width or bin number. The Freedman and Diaconis rule produces the best outlying aspects since it is a more robust rule.}.
The KDE employed in OAMiner and Beam uses the Gaussian kernel with the default bandwidth \citep{hardle1991}, as used by the authors \citep{Duan2015,Vinh2016} (stated in Section \ref{sec_related_work}). Euclidean distance was used in both OAMiner and Beam to compute the Gaussian kernel in the experiments. 

We conduct two sets of experiments in this section. First, we provide a runtime comparison between OAMiner, Beam, sGBeam and GBeam. The aim is to evaluate their relative runtime performances, as a means to compare the efficiency between the proposed density estimator and KDE used in these algorithms.
Second, we evaluate the quality of the subspaces discovered. This examines the quality of the subspaces discovered by these algorithms. 
Some algorithms produce a long list of subspaces. We examine up to the top 10 subspaces only when assessing the quality of each subspace.

We use six real-world datasets and five synthetic datasets which have diverse data characteristics in the experiments to examine the effects due to large data size and high number of dimensions.  
The data characteristics are given in Table \ref{tab:datasets}.
The synthetic datasets were used by previous studies \citep{Duan2015,Vinh2016} and   
 they were created by \citep{HiCS}. Each dataset has some outliers; each outlier stands out from normal clusters in at least 2 to 5 dimensions, and it may be an outlier in multiple subspaces independently.  Because these synthetic datasets have ground truths, they are used to assess the quality of the outlying aspects discovered by the miners. NBA was previous used by \citep{Duan2015}.
 ALOI is from the MultiView dataset collection (elki.dbs.ifi.lmu.de/wiki/DataSets/MultiView). Other datasets are from the UCI repository \citep{Lichman:2013}.

For each of the real-world datasets, a state-of-the-art anomaly detector called iForest \citep{Liu08,Emmott} was used to identify the top ten anomalies; and they were used as queries. Each outlying aspects miner identifies the outlying aspects for each query/anomaly. 

All outlying aspects miners conduct the search for subspaces with up to the maximum number of attributes $\ell=5$ for all datasets. The exceptions are Shuttle and P53Mutant which are used in the scaleup tests; here $\ell=3$ is used.

\begin{table}
 \centering
  \caption{Data characteristics of datasets. }
  \label{tab:datasets}
 \begin{tabular}{lrr}
 \toprule
 Dataset (DS)  & \#Attributes & Data size\\ 
 \midrule
Synthetic datasets & 10 - 50 & 1,000 \\
Shuttle &  10 & 49,097 \\
NBA & 20 & 220 \\
U2R & 34 & 60,821 \\
ALOI & 64 & 100,000 \\
Har & 551 & 5,272 \\
P53Mutant & 5,408 & 31,159 \\

 \bottomrule
 \end{tabular}
\end{table}

The machine used in the experiments has AMD 16-core CPU running at 2.3GHz with 64GB RAM; and the operating system is Ubuntu 15.04.

\subsection{Runtime comparison}
\label{sec:runtime-comparison}

The results comparing the runtimes of OAMiner, Beam, sGBeam and GBeam are presented in four groups of datasets: (1) Shuttle
and (2) P53Mutant for scaleup tests;  (3) four other real-world datasets; and (4) synthetic datasets.

\subsubsection{Scaleup test on increasing data sizes}
Figure \ref{fig_timing-a} shows the result on Shuttle.  sGBeam and GBeam are the fastest algorithms which have about the same runtimes. OAMiner failed to complete the task when the data size is more than 12000 (25\%) because its space complexity is quadratic to data size.

sGBeam and GBeam ran significantly faster than Beam and OAMiner, i.e., two orders of magnitude faster at 25\% data size. Note that sGBeam and GBeam at 100\% data size ran faster than OAMiner and Beam at 5\% data size! 

Beam is the slowest but it is able to complete the task using 100\% data size, albeit it is three orders of magnitude slower than sGBeam and GBeam.

Note that Beam should run faster than OAMiner because of its reduced search space. But OAMiner pre-computes the pairwise distances and stores them for table lookup in the actual run. The saving in time trades off for an increase in memory requirement. Thus, it limits OAMiner's applications to small datasets only, as shown in this experiment. OAMiner's reported runtime excluded the preprocessing time.

\subsubsection{Scaleup test on increasing dimensionality}

\begin{figure}
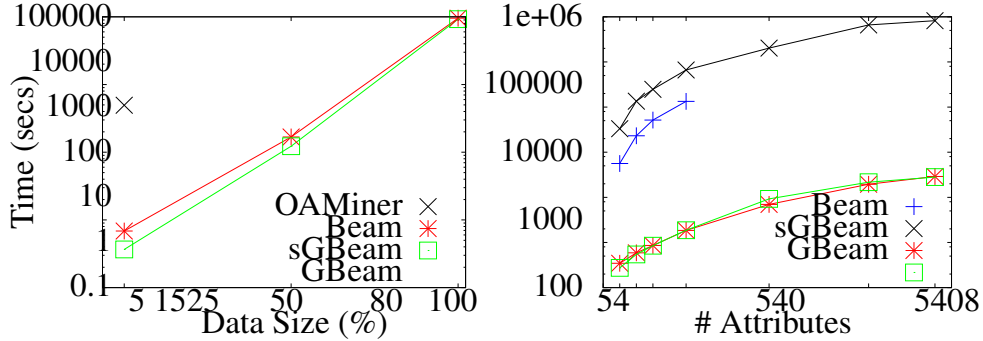

 \centering
 \subfloat[Runtime on Shuttle for 10 queries. Note that OAMiner failed to run using the data size of 50\% (24548 data points) on a machine having 64GB of RAM (`out of memory' error). \label{fig_timing-a}]{\includecombinedgraphics[scale=0.5,vecfile=./timing_attribute_size_total-eps-converted-to]{./timing_data_size_total}}%
 \subfloat[Runtime on P53Mutant for a single query.  OAMiner failed to run due to `out of memory' error. Beam was unable to complete the tasks with higher dimensions within 12 days.\label{fig_timing-b}]{\includecombinedgraphics[scale=0.5,vecfile=./timing_data_size_total-eps-converted-to]{./timing_attribute_size_total}}
 
 \caption{Scaleup tests using $\ell=3$: (a) Increasing data size; (b) increasing number of dimensions. }
\end{figure}

The scaleup test on P53Mutant, as the number of attributes increases from 54 to 5408, is showed in Figure \ref{fig_timing-b}. Only sGBeam and GBeam were able to complete the scaleup test. Beam could only complete the task with 54 attributes within a reasonable time. Beam is estimated to take more than 100 days to complete the task with 540 attributes; and a number of years for the task with 5408 attributes. OAMiner was unable to run due to out of memory error even with 54 attributes.

\begin{table}
  \centering
  \caption{Runtime result  (seconds) for 10 queries on each real-world dataset. %
  $\ell=5$.
  OM denotes `Out of Memory' error.} %
  \begin{tabular}{lrrrr}
    \toprule
DS &  OAMiner & Beam & sGBeam & GBeam\\
    \midrule
NBA & 5 & 116 & 1.25 & 1.16 \\
U2R & OM & $>$12 Days & 4,212 & 4,170 \\
ALOI &  OM & $>$12 Days & 52,828 & 47,810 \\
Har &  OM & $>$12 Days & 5,410 & 4,725 \\
\bottomrule
  \end{tabular}
  \label{time_other_datasets}
\end{table}

\subsubsection{Other real-world datasets} 
Table \ref{time_other_datasets} shows the runtimes on four real-world datasets. Again, only sGbeam and GBeam were able to complete all the tasks. OAMiner had out of memory errors in three large datasets; and Beam could not complete the same three datasets within 12 days. Beam is three orders of magnitude slower than sGbeam and GBeam on the NBA dataset. 

\begin{table}
  \centering
  \caption{Runtime results for the synthetic datasets (seconds).}  %
  \vspace{-2mm}
  \begin{tabular}{ccrrrr}
    \toprule
DS & \#Query & OAMiner & Beam & sGBeam & GBeam\\
    \midrule
10D & 19 & 35 & 986 & 1.3 & 1.2 \\
20D & 25 & 1,187 & 10,242 & 11 & 9 \\
30D & 44 & 16,625 & 26,230 & 39 & 24  \\
40D & 53 & 102,457 & 52,490 & 75 & 46 \\
50D & 68 & 451,727 & 90,075 & 143 & 79 \\
    \bottomrule
  \end{tabular}
  \label{time_synthetic}
    \vspace{-2mm}
\end{table}

\subsubsection{Synthetic datasets}
Table \ref{time_synthetic} shows the runtimes on the five synthetic datasets. sGBeam and GBeam are the fastest algorithms\footnote{Note that the differences in runtime between sGBeam and GBeam are due to difference in the total number of subspaces searched for all queries. 
The algorithm is made efficient as follows: The computed density z-scores for all explored subspaces for the previous queries are saved to be used as a lookup for subsequent queries when the same subspaces need to be explored again. %
The proposed density estimator used enables sGBeam to explore more distinct subspaces than those explored by GBeam because of sGBeam's smoothed estimations.  For example on 50D, sGBeam explored nearly twice the number of subspaces than GBeam. Despite this, both sGBeam and GBeam have the same order of runtime.}; they ran three orders of magnitude faster than Beam, and four orders of magnitude faster than OAMiner on the 50D dataset.

\noindent
{\bf Summary}.
The runtime comparison shows that the proposed efficient density estimator enables the beam search method to run two to three orders of magnitude faster than that using KDE, enabling the method to deal with large datasets with thousands of dimensions that would otherwise be impossible.
The use of KDE slows down OAMiner and Beam significantly in comparison with sGBeam and GBeam.

\subsection{Qualitative evaluation}
\label{sec:OAMinerDatasets}

Here we focus on assessing the quality of the subspaces discovered. We use  the synthetic datasets, and the three real-world datasets which  OAMiner and Beam could run in the last section. They are described in two subsections below.

\noindent
{\bf Synthetic datasets}. Table \ref{tab:summary} provides the summary results of the  five datasets having 10 to 50 dimensions.
In each algorithm, we report two qualitative results, i.e., the number of exact matches with the ground truths (shown in parenthesis), and the number of matches which includes subsets or supersets of ground truths, in addition to exact matches. 

In terms of the number of exact matches, Beam\footnote{Note that the Matlab code of Beam \citep{Vinh2016} produced worse than the result presented here with an average of 179.5 (98).} and sGBeam are the best performers which produced a total of 149.5 and 144 exact matches, respectively. Both results are about 50\% more than those obtained by GBeam (97). 
OAMiner  with only 13.5 exact matches is the worst performer.

In terms of the number of matches (which include either subsets or supersets of the ground truths), Beam and GBeam produce approximately the same result (190.5 and 198); followed by sGBeam (179.5) and OAMiner (168).

\begin{table}
 \centering
  \caption{Summary results of five synthetic datasets. Each algorithm shows two results: Number within the parenthesis is the number of exact matches; and the number of matches which includes the subsets and supersets of ground truths in addition to the exact matches. A match with one of the two subspaces that exist in a ground truth is awarded with 0.5.} %
 \label{tab:summary}
\sisetup{
  input-symbols         = (),
  table-format          = 3.1,
} 
\setlength{\tabcolsep}{2pt} %
 \begin{tabular}{@{}crrSrSrSr}
  \toprule

DS & \multicolumn{2}{c}{OAMiner} & \multicolumn{2}{c}{Beam}  & \multicolumn{2}{c}{sGBeam} & \multicolumn{2}{c}{GBeam}\\
  \midrule
10D & 13 & (1.5) & 19 & (19)\hspace{2.6mm} & 19 & (19)\hspace{2.6mm} & 19 & (14)\hspace{2.6mm} \\
20D & 22 & (1)\hspace{2.6mm} & 24 & (10)\hspace{2.6mm} & 20 & (10)\hspace{2.6mm} & 25 & (10)\hspace{2.6mm} \\
30D & 37 & (3)\hspace{2.6mm} & 42.5 & (38.5)  & 41.5 & (35.5) & 43.5 & (19.5) \\
40D & 44 & (4)\hspace{2.6mm} & 47.5 & (32.5) & 39 & (31)\hspace{2.6mm} & 49 & (24)\hspace{2.6mm}  \\
50D & 52 & (4)\hspace{2.6mm} & 57.5 & (49.5)  & 55.5 & (48.5) & 61.5 & (29.5) \\
  \midrule
Total & 168	& (13.5) &	190.5	& (149.5) &	175 & 	(144)\hspace{2.6mm} & 198 & (97)\hspace{2.6mm} \\

  \bottomrule
 \end{tabular}
\vspace{7mm}
 \centering
  \caption{Examples of matches and exact matches on the 10-dimension synthetic dataset. %
  Discovered subspaces which have exact matches with ground truths are bold-faced. ID is the instance id of a query; GT is the Ground Truth. The numbers in a bracket (i.e., subspace) are attribute indices.}
   \label{tab:synth-10D}
 \begin{tabular}{@{}r@{$\ $}L{1.35cm}@{$\ $}L{2cm}@{$\ $}L{1.35cm}@{$\ $}L{1.35cm}@{$\ $}L{1.35cm}@{}}
  \toprule
ID & GT &  {OAMiner} & {Beam} & {sGBeam} & {GBeam} \\
  \midrule
220 & \{2,3,4,5\} & \{4\} \{2\} \{3\} & \{\bf 2,3,4,5\} & \{\bf 2,3,4,5\} & \{7,8,6,9\} \{3,4,5\}  \\
315 & \{0,1\} \{6,7\} &  \{6\} \{4\} \{3\} \{7\} \{\bf 0,1\} & \{\bf 0,1\} \{\bf 6,7\} & \{\bf 0,1\} \{\bf 6,7\} & \{\bf 0,1\} \{\bf 6,7\} \\
323 & \{8,9\} & \{6\} \{2,8,9\} & \{\bf 8,9\} & \{\bf 8,9\} & \{\bf 8,9\} \\
577 & \{2,3,4,5\} &  \{0\} \{4\} & \{6,7\} \{\bf 2,3,4,5\} & \{\bf 2,3,4,5\} & \{3,4,5\}  \\
723 & \{2,3,4,5\} & \{0\} \{3\} \{9\} & \{\bf 2,3,4,5\} & \{\bf 2,3,4,5\} & \{2,4\} \\
  \bottomrule
 \end{tabular}
\end{table}

Table \ref{tab:synth-10D} shows examples of the subspaces found by OAMiner, Beam, sGBeam and GBeam on the synthetic dataset having 10 dimensions\footnote{Note that the OAMiner result is different from that presented by \citep{Duan2015} because they have reported to have eliminated all anomalies in the dataset, except the anomaly currently under investigation. Our run includes all anomalies, which is more realistic in practice, as the information of all anomalies is usually unavailable, even with the aid of an anomaly detector. The presented results in both Table \ref{tab:summary} and Table \ref{tab:synth-10D} were obtained by using the original datasets, rather than the reduced datasets.}.

OAMiner employs a search pruning rule which relies on minimal subspaces. That has prevented OAMiner from finding the ground truth subspaces which have a high number of attributes, as it stops searching when subspaces with low dimensionality have been found.
GBeam has an issue in discovering ground truth subspaces with four or more attributes. This is because the estimated density is not accurate enough. Both sGBeam and Beam have no such issue.

\noindent
{\bf Real-world datasets}. Here we focus on Beam and sGBeam to examine how similar their discovered subspaces are.  The results from three queries in each of NBA, Shuttle and ALOI\footnote{We used 108 points of one cluster only (out of 900 clusters) in ALOI for this experiment. Otherwise Beam could not complete within a reasonable time.} are shown in Table \ref{tab:my_label}. The results show that the subspaces discovered by Beam and sGBeam are similar; only a few subspaces found by one algorithm are not in the top ten subspaces discovered by the other algorithm in two queries.

\begin{table}
 \centering
 \caption{Subspaces discovered by Beam and sGBeam on NBA, Shuttle and ALOI (in the first, second and third rows, respectively). The top three subspaces found for each of three queries are list. $^*$ indicates the subspace which does not appear in the top ten list discovered by the other algorithm.  ID is the instance id of a query; the numbers in a bracket (i.e., subspace) are attribute indices.}
 \label{tab:my_label}
 \setlength{\tabcolsep}{2pt} %
\begin{tabular}{cll}
  \toprule
ID & Beam & sGBeam \\
  \midrule
0 & \{3\} \{2\} \{19\} & \{1,3\} \{3\} \{1,19\} \\1 & \{19\} \{1,19\} \{6\} & \{1,19\} \{19\} \{6\} \\
157 & \{1,2\} \{11\}$^*$ \{2,3\} & \{1,3,19\} \{1,2,3\} \{1,19\}$^*$ \\
  \midrule
24147 & \{0,7\} \{0,4\} \{0,4,8\} & \{0,4\} \{0,8\} \{0,7\} \\
35414 & \{0,7\} \{0,4\} \{0,4,8\} & \{0,4\} \{0,7\} \{0,8\} \\
37048 & \{0,7\} \{0,4\} \{0,4,8\}  & \{0,4\} \{0,7\} \{0,8\} \\  \midrule
13 & \{16,36\} \{0,36\} \{16,61\}$^*$ & \{0,16,36\} \{0,36\} \{16,36\} \\
24 & \{0,16\} \{36\} \{16,36\} & \{0,16\} \{36\} \{0,16,36\} \\

28 & \{21\} \{16,61\} \{16,25\} & \{21\} \{0,16,61\} \{0,61\} \\  \bottomrule
 \end{tabular}
 \end{table}

Note that we do not have the ground truths to verify the subspaces discovered on the real-world datasets.
Without ground truths, there are no suitable summary measures for qualitative assessment of outlying aspects mining. Consensus Index, which is based on entropy, was used previously \citep{Vinh2016}; but it is a measure of consensus of (characteristic) features discovered  for instances of a cluster. It is more suitable for measuring clustering results than the `correctness' of distinguishing features of an anomaly, which is the aim of an outlying aspects miner, where different anomalies are likely to have different distinguishing features even with reference to the same cluster.

\section{Conclusions}

This paper shows that a systematic search method runs orders of magnitude faster by simply replacing the commonly used density estimator with the {\bf sGrid} density estimator. This enables  the systematic search method to handle large datasets with thousands of attributes, an impossible task for existing implementations such as OAMiner \citep{Duan2015} and Beam \citep{Vinh2016} that use KDE. We show that sGBeam runs at least two orders of magnitude faster than Beam and  OAMiner in our experiments. This is achieved with a minor degradation in terms of the number of exact matches found compared with Beam that uses KDE---a small price to pay for the huge gain in runtime.

Compared with the ordinary grid-based density estimator, {\bf sGrid} which is a smoothed version overcomes its known weakness to produce more accurate estimations.  This enables sGBeam to discover 50\% more exact matches than GBeam on the synthetic datasets, while still maintaining the same order of runtime.

{\bf sGrid} is a generic estimator and can be used in all existing algorithms that employ density estimators. Simply replacing the computationally expensive density estimators with {\bf sGrid} will significantly improve their runtimes.

\bibliographystyle{named}
\bibliography{stout-ref}

\end{document}